\documentclass[11pt]{article}

\PassOptionsToPackage{sort}{natbib}
\usepackage[final]{acl}

\usepackage{booktabs}
\usepackage{multirow}
\usepackage{enumitem}

\usepackage[utf8]{inputenc}
\DeclareUnicodeCharacter{2212}{-}

\definecolor{authority}{HTML}{6A4C93}
\definecolor{care}{HTML}{00B4A6}
\definecolor{fairness}{HTML}{3498DB}
\definecolor{loyalty}{HTML}{E74C3C}
\definecolor{sanctity}{HTML}{F39C12}

\usepackage{times}
\usepackage{latexsym}

\usepackage{algorithm}
\usepackage{algpseudocode}
\usepackage{listings}
\usepackage{tikz}
\usetikzlibrary{shapes,arrows,fit,positioning}
\usepackage{tcolorbox}

\usepackage{orcidlink}

\usepackage[T1]{fontenc}

\usepackage[utf8]{inputenc}

\usepackage{microtype}

\usepackage{caption}
\usepackage{subcaption}
\usepackage{svg}

\usepackage{inconsolata}

\usepackage{graphicx}

\usepackage{amsmath}

\usepackage[nameinlink]{cleveref}

\title{Beyond Human Judgment: A Bayesian Evaluation of LLMs' Moral Values Understanding}

\author{Maciej Skorski\ \orcidlink{0000-0003-2997-7539} \\
  University of Luxembourg \\
  \texttt{maciej.skorski@gmail.com} \\\And
  Alina Landowska\ \orcidlink{0000-0002-7966-8243} \\
  Kozminski University \and SWPS University \\
  \texttt{alandowska@kozminski.edu.pl} \\}

\begin{document}

\maketitle

\begin{abstract}
How do Large Language Models understand moral dimensions compared to humans?

This first comprehensive large-scale Bayesian evaluation of leading language models provides the answer. In contrast to prior approaches based on deterministic ground truth (obtained via majority or inclusion consensus), we obtain the labels by modelling annotators' disagreement to capture both aleatoric uncertainty (inherent human disagreement) and epistemic uncertainty (model domain sensitivity).

We evaluated Claude Sonnet 4, DeepSeek-V3, and Llama 4 Maverick across 250K+ annotations from nearly 700 annotators in 100K+ texts spanning social networks, news, and discussion forums. Our GPU-optimized Bayesian framework processed 1M+ model queries, revealing that AI models generally rank among the top 25\% of annotators in terms of balanced accuracy, substantially better than average humans.

Importantly, we find that AI produces far fewer false negatives than humans, highlighting their sensitive moral detection capabilities.

\textbf{Keywords:}  Computational Ethics, Large Language Models,  Moral Foundation Theory, Bayesian modeling, soft labels
\end{abstract}

\noindent\textbf{Extended version:} \href{https://arxiv.org/abs/2508.13804}{arXiv:2508.13804} \\
\noindent\textbf{Supplementary materials:} \href{https://osf.io/tpzau/}{osf.io/tpzau}

\section{Introduction}

\subsection{Background}

Moral Foundations Theory (MFT) provides a comprehensive framework for understanding human moral reasoning across cultures, identifying core dimensions typically expressed as virtue/vice pairs: Care vs.\ Harm, Fairness vs.\ Cheating, Loyalty vs.\ Betrayal, Authority vs.\ Subversion, and Sanctity vs.\ Degradation \citep{graham2013moral, haidt2012righteous}. These foundations shape individual and collective decision-making, from political preferences to social behavior \citep{feinberg2013moral,graham2009liberals,royAnalysisNuancedStances2021,nguyen2022mapping}, making their computational detection crucial for understanding discourse dynamics and developing ethically-aligned AI systems.

\begin{table}[h]
\centering
\resizebox{0.99\columnwidth}{!}{
\begin{tabular}{p{0.79\columnwidth} p{0.20\columnwidth}}
\toprule
\textbf{Text} & \textbf{Foundation} \\
\midrule
"My heart breaks seeing children separated from families at the border" & \textcolor{care}{\textbf{Care}} \\
"Everyone deserves equal access to healthcare regardless of income" & \textcolor{fairness}{\textbf{Fairness}} \\
"Respect your elders and follow traditional values that built this nation" & \textcolor{authority}{\textbf{Authority}} \\
"Stand with our troops - they sacrifice everything for our freedom" & \textcolor{loyalty}{\textbf{Loyalty}} \\
"Marriage is sacred and should be protected from secular corruption" & \textcolor{sanctity}{\textbf{Sanctity}} \\
\bottomrule
\end{tabular}
}
\caption{Posts and Associated Moral Foundation}
\end{table}

The computational linguistics community has successfully fine-tuned pre-trained language models to predict moral values \citep{nguyenMeasuringMoralDimensions2024a,zangari-etal-2025-me2,preniqiMoralBERTFineTunedLanguage2024}, achieving good alignment with human judgment when domain similarity and sufficient training data are available. However, systematic evaluation of large language models (LLMs) remains limited despite their rapid advances and potential as a compelling alternative that should suffer less from poor generalization and distribution shift.

This paper addresses these limitations through rigorous large-scale evaluation of state-of-the-art language models across established moral reasoning corpora, employing Bayesian methods to resolve disagreeing annotations.

\subsection{Contribution}

\textbf{Bayesian uncertainty modelling of moral annotations.} We introduce Bayesian modelling of annotator disagreements for moral foundation evaluation, moving beyond simple deterministic ground-truth assumptions. This captures both aleatoric uncertainty (inherent human disagreement) and epistemic uncertainty (model sensitivity across domains and foundations).

\textbf{Large-scale evaluation.} We analyse market-leading large language models with 1M+ queries across 100K+ texts and 250K+ annotations from diverse sources, providing the most comprehensive moral reasoning evaluation to date.

\textbf{Statistical analysis of Type I/II errors.} We demonstrate that AI performs comparably to top annotators in balanced accuracy, considerably improving false negatives at the price of slightly increased false positive rates - contrary to fears that AI may underpredict moral values.

\textbf{Novel GPU-optimized implementation of Bayesian labels.} We developed a  TensorFlow framework using sparse operations for scalable Bayesian inference on soft labels, of interest to the computational linguistics community.

\subsection{Related Work}

\textbf{Dictionary methods} for predicting moral values were initially developed~\cite{hoppExtendedMoralFoundations2021} and remain widely used in applied studies, but demonstrate poor precision compared to pre-trained language models~\cite{nguyenMeasuringMoralDimensions2024a}.

\textbf{Pretrained language models} show strong alignment with human moral judgments given sufficient training data~\cite{nguyenMeasuringMoralDimensions2024a,zangari-etal-2025-me2,preniqiMoralBERTFineTunedLanguage2024}, though they suffer from distribution shift and poor cross-domain generalization. This makes large language models a compelling alternative.

\textbf{Large language models} were recently compared to human performance~\cite{bullaLargeLanguageModels2025}, concluding LLMs superiority. However, their strict majority voting for ground truth excluded nuanced moral content, retaining only less ambiguous cases and departing from the inclusive consensus practices (marked when flagged by at least one annotator)~\cite{nguyenMeasuringMoralDimensions2024a,zangari-etal-2025-me2,preniqiMoralBERTFineTunedLanguage2024}. Additionally, leave-one-out estimation of annotator ranking lacked statistical robustness. Another recent work~\cite{skorskiMoralGapLarge2025} found LLMs showing imbalance - lower than anticipated recall - under inclusive annotator consensus.

Our Bayesian method resolves these problems by modeling annotator disagreement, striking the balance between inclusive and overly exclusive consensus rules, while our large-scale evaluation across multiple datasets and modern models ensures robustness of findings.

\textbf{More on moral foundation theory.} For comprehensive coverage of (computational) moral foundation theory challenges, see~\cite{zangariSurveyMoralFoundation2025}.

\section{Data and Methods}

\subsection{Datasets}

Our robust evaluation utilizes three established and diverse corpora totaling 250K+ annotations of moral values from hundreds of annotators with diverse expertise (from experts to crowd-workers) across 100K+ texts spanning social media, news, and forum discussions.

\textbf{Moral Foundations Twitter Corpus (MFTC)} \cite{hooverMoralFoundationsTwitter2020}: 128,454 annotations from 23 trained annotators across 33,686 tweets from seven discourse domains.

\textbf{Extended Moral Foundations Dictionary (eMFD)} \cite{hoppExtendedMoralFoundations2021}: 73,001 crowd-sourced annotations from 654 contributors on 54,867 text segments extracted from approximately 8,000 news documents by major outlets, including The Washington Post, CNN, Fox News selected via GDELT.

\textbf{Moral Foundations Reddit Corpus (MFRC)} \cite{tragerMoralFoundationsReddit2022}: 61,226 annotations from 6 trained coders across 17,885 Reddit comments from 12 subreddits covering politics and everyday moral discourse.

\subsection{Bayesian Annotation Competence Model}

As in the related work, we frame the problem as binary prediction tasks for individual moral foundations (abbreviated to care, fairness, authority, loyalty, sanctity). Optionally, we consider the aggregated "any moral content" category derived from positive labels across foundations. We note that this aggregated category is inherently unreliable when only positive labels are provided (as in eMFD), since true negatives cannot be distinguished from unlabeled moral content, creating systematic classification ambiguity. 

Given substantial inter-annotator disagreement demonsrated by PABAK scores in \Cref{tab:pabak}, we resort to probabilistic (Bayesian) methods of obtaining ground-truth labels~\cite{paunAggregatingLearningMultiple2021}.

\begin{table}[h!]
\centering
\begin{tabular}{@{}lrrr@{}}
\toprule
Foundation & MFTC & MFRC & eMFD \\
\midrule
Care & 0.71 & 0.67 & 0.33 \\
Fairness & 0.63 & 0.64 & 0.36 \\
Loyalty & 0.62 & 0.82 & 0.38 \\
Authority & 0.52 & 0.78 & 0.41 \\
Sanctity & 0.58 & 0.83 & 0.44 \\
Any & 0.34 & 0.38 & 1.00 \\
\bottomrule
\end{tabular}
\caption{Prevalence-Adjusted Bias-Adjusted Kappa (PABAK) scores measuring inter-annotator agreement across datasets and moral foundations}
\label{tab:pabak}
\end{table}

We model annotator disagreements using a variant of Dawid-Skene's model~\cite{dawidMaximumLikelihoodEstimation1979} with weak Dirichlet priors to estimate ground-truth labels and annotator reliability:

\textbf{Model Specification.} We assume $N$ texts, $J$ annotators, and $K$ categories. True category prevalences follow $\boldsymbol{\pi} \sim \text{Dir}(\boldsymbol{\alpha})$ where $\boldsymbol{\alpha} = (1, 1, ..., 1)$ provides uniform priors. Each annotator $j$ has a confusion matrix $\boldsymbol{\Theta}_j$ with rows $\boldsymbol{\theta}_{jk} \sim \text{Dir}(\boldsymbol{\beta}_k)$, where $\boldsymbol{\beta}_k$ is a $K$-dimensional vector with $\beta_{kk} = 2$ (diagonal) and $\beta_{kl} = 0.5$ for $l \neq k$, encoding weak belief that annotators correctly identify majority of categories.
For text $i$ with true category $z_i$ and annotations $\mathbf{y}_i = (y_{i1}, ..., y_{iJ})$ we have:
$$\Pr\{z_i = k \mid \mathbf{y}_i, \boldsymbol{\pi}, \boldsymbol{\Theta}\} \propto \pi_k \prod_{j=1}^J \theta_{jk,y_{ij}},$$
This posterior accounts for varying annotator reliability while estimating both confusion matrices and ground-truth labels. For computational stability and reliable convergence, all calculations are performed in the log-domain using logits.

\begin{figure}[htbp]
\centering
\resizebox{0.75\linewidth}{!}{
\begin{tikzpicture}[
    obs/.style={draw,fill=gray!30,rectangle,minimum size=8mm},
    latent/.style={draw,fill=gray!10,circle,minimum size=8mm},
    param/.style={draw,circle,minimum size=8mm},
    hyper/.style={draw,rectangle,fill=blue!20,minimum size=6mm},
    ->
]

\node[hyper] (alpha) at (0,3) {$\boldsymbol{\alpha}$};
\node[hyper] (beta) at (6,3) {$\boldsymbol{\beta}$};

\node[param] (pi) at (1,2) {$\boldsymbol{\pi}$};
\node[param] (theta) at (5,2) {$\boldsymbol{\Theta}_j$};

\node[latent] (z) at (2,0.5) {$z_i$};

\node[obs] (y) at (4,0.5) {$y_{ij}$};

\draw[thick,rounded corners=3pt] (1.5,-0.3) rectangle (4.5,1.3);
\node at (1.7,1.1) {\small $N$};

\draw[thick,rounded corners=3pt] (3.5,-0.3) rectangle (5.5,2.7);
\node at (3.7,2.5) {\small $J$};

\draw[->] (alpha) -- (pi);
\draw[->] (beta) -- (theta);
\draw[->] (pi) -- (z);
\draw[->] (z) -- (y);
\draw[->] (theta) -- (y);

\end{tikzpicture}
}
\caption{Graphical model representation of the model for multi-annotator classification. Light gray circles represent latent variables, dark gray rectangles represent observed variables, white circles represent parameters, and blue rectangles represent hyperparameters. Plates indicate replication over items ($N$) and annotators ($J$).}
\label{fig:dawid_skene}
\end{figure}

\textbf{Implementation.} We developed a GPU-optimized implementation in TensorFlow~\cite{tensorflow2015-whitepaper} using custom graphs and sparse tensor operations for computational efficiency. The framework supports MAP estimation and Monte Carlo sampling with optional fixed effects modeling (\Cref{sec:implementation}). Testing on A100 GPUs achieved 1,000 optimization steps per second for ~100k annotations, enabling rapid convergence across large-scale datasets.

\subsection{Large Language Models}

Three recent advanced large language models were analysed for moral foundations classification: (1) \textbf{Claude Sonnet 4} (Anthropic, 2025) with a 200,000 token context window; (2) \textbf{DeepSeek-V3} (DeepSeek, 2024) with 671 billion total parameters, 37 billion active parameters, and a 128,000 token context window; and (3) \textbf{Llama 4 Maverick} (Meta, 2025) with 402 billion total and 17 billion active parameters, and a 256,000 token context window.

All models were queried using the temperature setting of 0.30 to balance deterministic responses with modest variability. Each text was analysed individually (no batching of inputs) to ensure focused classification results. The prompt used was:

\begin{tcolorbox}[
    colback=blue!5,
    colframe=blue!50!black,
    boxrule=1.5pt,
    arc=4pt,
    left=1pt,
    right=1pt,
    top=5pt,
    bottom=5pt,
    title={\textbf{Moral Foundations Theory Classification}},
    fonttitle=\bfseries,
    coltitle=white,
    colbacktitle=blue!75!black
]

{\small You are an expert in moral psychology, classifying text according to Haidt's theory.

For each moral foundation, mark \texttt{true} if moral values from that foundation are expressed in the text, \texttt{false} if not expressed.

\vspace{5pt}

\textbf{Answer only with a valid JSON in this format:}

\begin{verbatim}
{
  "care/harm": [true / false],
  "fairness/cheating": [true / false], 
  "loyalty/betrayal": [true / false],
  "authority/subversion": [true / false],
  "sanctity/degradation": [true / false]
}
\end{verbatim}
}

\end{tcolorbox}

Experiments were tracked using W\&B\footnote{\url{https://wandb.ai/}} and repeated to ensure stability and reproducibility. 

\Cref{sec:prompt} provides details and rationale about the prompt selection.

\subsection{Metrics}

We apply standard classification metrics to Bayesian-estimated confusion matrices: {Balanced Accuracy} $\frac{1}{2}(\frac{TP}{TP+FN} + \frac{TN}{TN+FP})$, {Precision} $\frac{TP}{TP+FP}$, {Recall} $\frac{TP}{TP+FN}$, {False Positive Rate} $\frac{FP}{FP+TN}$, and {False Negative Rate} $\frac{FN}{FN+TP}$.

\subsection{Supplementary Materials}

Code, data preprocessing scripts, experimental results, and additional evaluation results are available from the project repository~\cite{skorskiBayesianEvaluationLLMs2025}.

\section{Results}

Bayesian models were fitted for each language model and dataset, the model considered a one additional annotator each time, and confusion matrices obtained through MAP were used to calculate classification metrics and percentile ranks.

\subsection{Accuracy Analysis}

\begin{figure}[h!]
   \centering
   \includegraphics[width=0.95\linewidth]{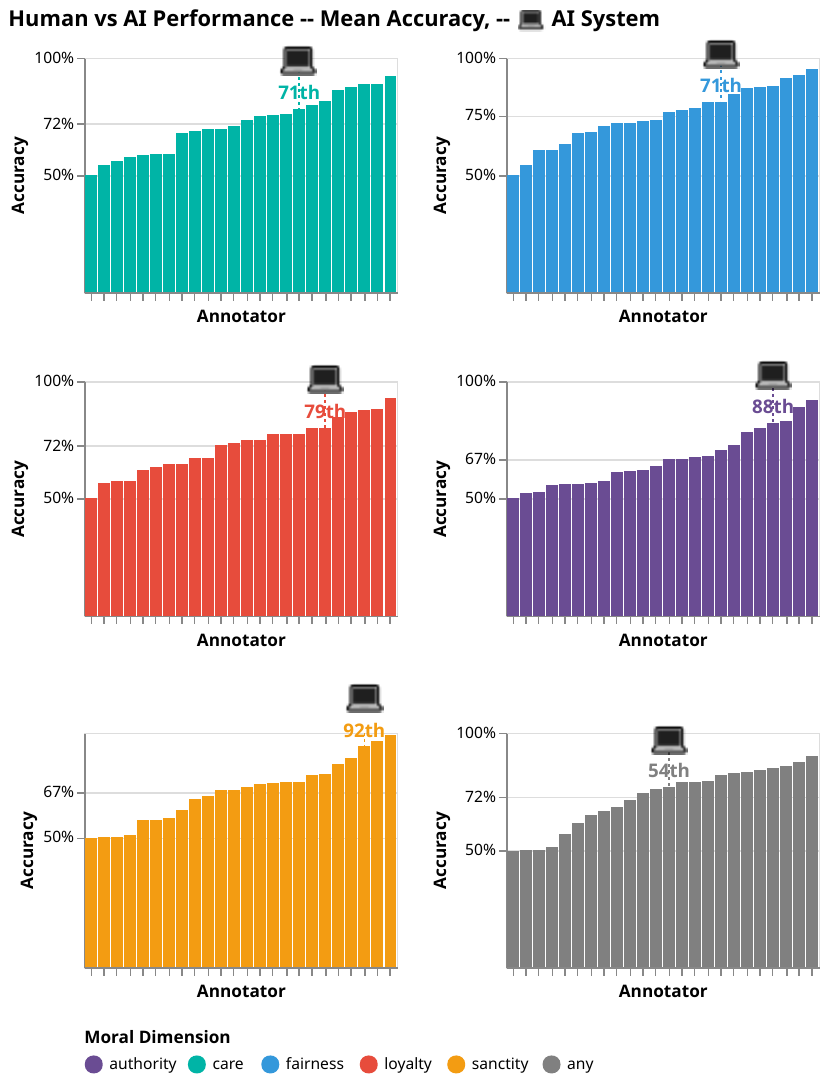}
   \caption{DeepSeek-V3 vs human accuracy (MFTC).}
   \label{fig:mftc-accuracy}
\end{figure}

\begin{figure}[h!]
   \centering
   \includegraphics[width=0.95\linewidth]{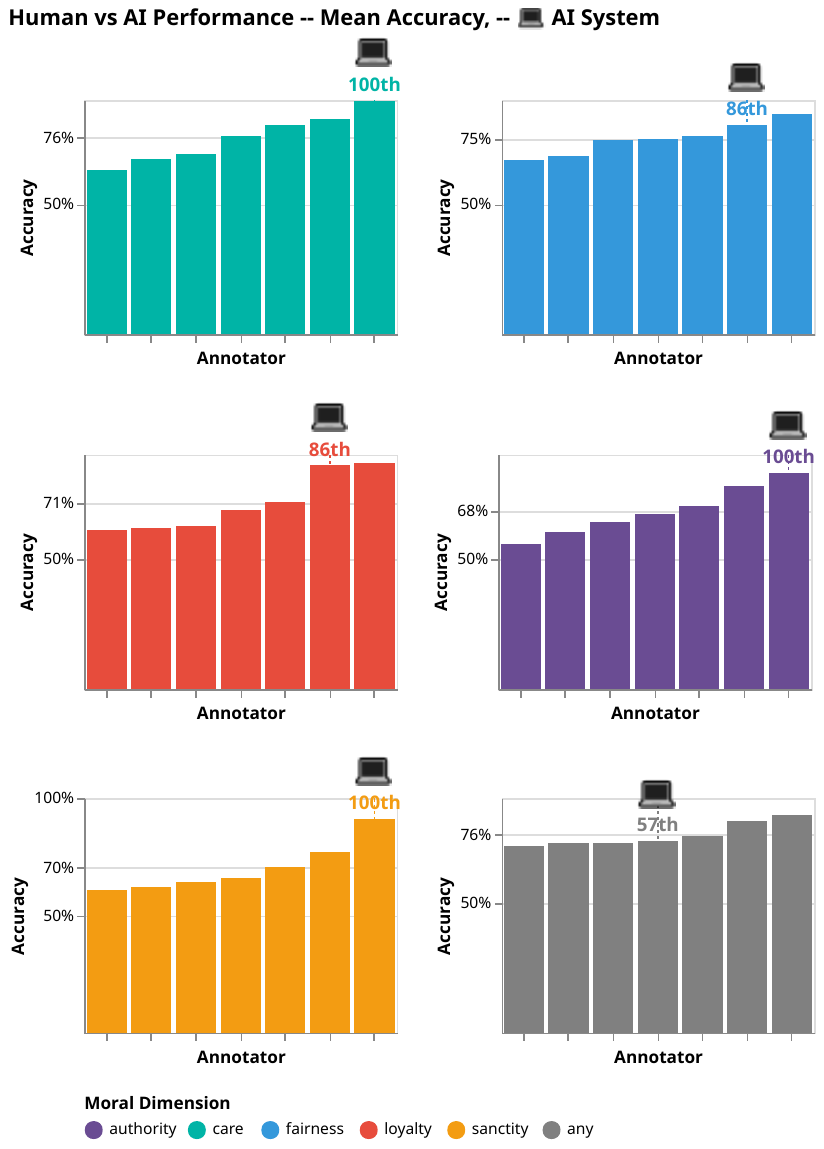}
   \caption{Claude Sonnet 4 vs human accuracy (MFRC).}  
   \label{fig:mfrc-accuracy}
\end{figure}

\begin{figure}[h!]
   \centering
   \includegraphics[width=0.95\linewidth]{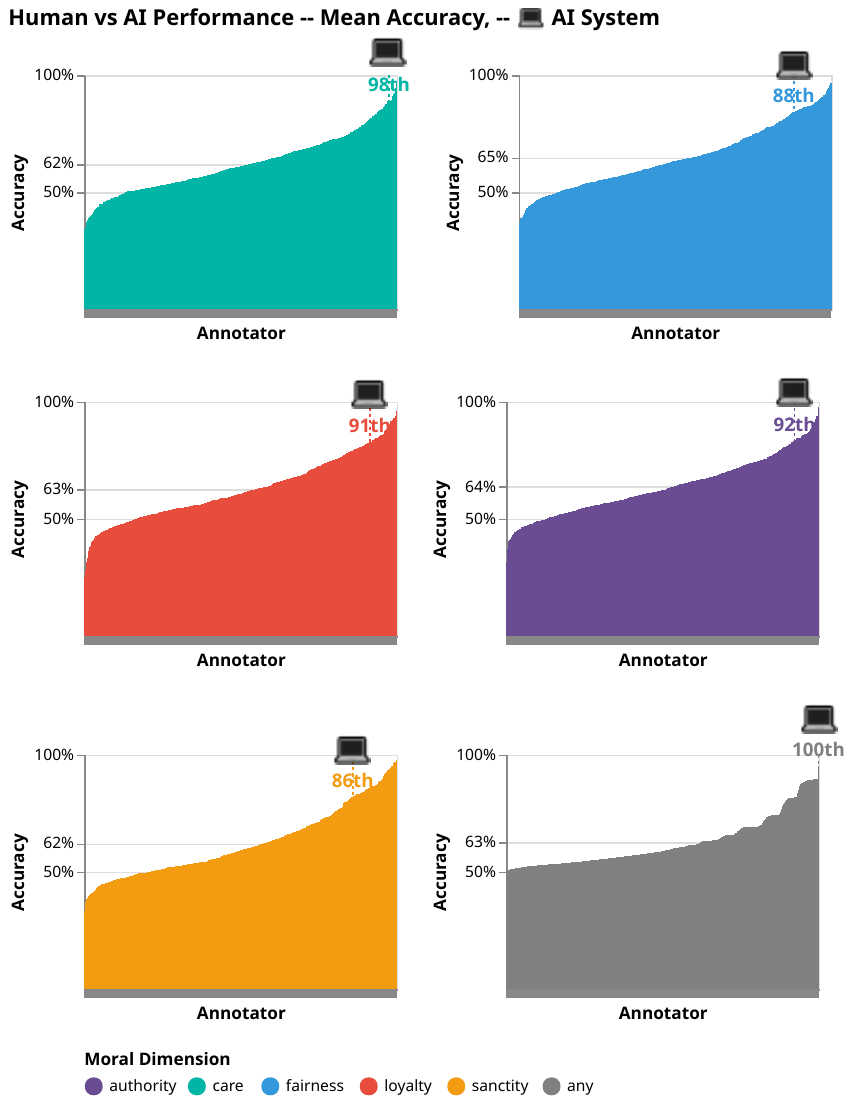}
   \caption{Llama 4 Maverick vs humans (eMFD).}
   \label{fig:emfd-accuracy}
\end{figure}

Balanced accuracy results appear in \Cref{tab:accuracy} and \Cref{fig:mftc-accuracy,fig:mfrc-accuracy,fig:emfd-accuracy}.

\begin{table}[h!]
   \centering
   \resizebox{0.96\linewidth}{!}{
\begin{tabular}{@{}llrrrrrr@{}}
\toprule
\multicolumn{2}{c}{} & \multicolumn{6}{c}{Moral Foundations} \\
\cmidrule(lr){3-8}
Model & Metric & Any & Authority & Care & Fairness & Loyalty & Sanctity \\
\midrule
\multicolumn{8}{c}{\textbf{MFTC Dataset}} \\
\midrule
\multirow{2}{*}{Claude Sonnet 4} & Acc\% & 75 & 80 & 78 & 82 & 81 & 90 \\
& Pct & 50 & 83 & 75 & 71 & 79 & 100 \\
\addlinespace[0.3em]
\multirow{2}{*}{DeepSeek-V3} & Acc\% & 77 & 82 & 78 & 81 & 80 & 85 \\
& Pct & 54 & 88 & 71 & 71 & 79 & 92 \\
\addlinespace[0.3em]
\multirow{2}{*}{Llama 4 Maverick} & Acc\% & 68 & 79 & 76 & 83 & 80 & 87 \\
& Pct & 42 & 79 & 63 & 71 & 75 & 96 \\
\addlinespace[0.3em]
Human & Avg\% & 72 & 67 & 71 & 75 & 72 & 67 \\
\midrule
\multicolumn{8}{c}{\textbf{eMFD Dataset}} \\
\midrule
\multirow{2}{*}{Claude Sonnet 4} & Acc\% & 85 & 74 & 88 & 83 & 72 & 80 \\
& Pct & 95 & 75 & 97 & 91 & 78 & 84 \\
\addlinespace[0.3em]
\multirow{2}{*}{DeepSeek-V3} & Acc\% & 93 & 81 & 88 & 82 & 85 & 75 \\
& Pct & 100 & 89 & 98 & 85 & 93 & 77 \\
\addlinespace[0.3em]
\multirow{2}{*}{Llama 4 Maverick} & Acc\% & 95 & 83 & 89 & 84 & 83 & 82 \\
& Pct & 100 & 92 & 98 & 88 & 91 & 86 \\
\multirow{2}{*}{GPT-5 mini} & Acc\% & 82 & 64 & 78 & 77 & 62 & 66 \\
 & Pct & 92 & 46 & 81 & 76 & 45 & 49 \\
\addlinespace[0.3em]
Human & Avg Acc\% & 63 & 64 & 62 & 64 & 62 & 63 \\
\midrule
\multicolumn{8}{c}{\textbf{MFRC Dataset}} \\
\midrule
\multirow{2}{*}{Claude Sonnet 4} & Acc\% & 74 & 83 & 90 & 80 & 86 & 91 \\
& Pct & 57 & 100 & 100 & 86 & 86 & 100 \\
\addlinespace[0.3em]
\multirow{2}{*}{DeepSeek-V3} & Acc\% & 68 & 83 & 90 & 73 & 84 & 82 \\
& Pct & 14 & 86 & 100 & 43 & 86 & 100 \\
\addlinespace[0.3em]
\multirow{2}{*}{Llama 4 Maverick} & Acc\% & 62 & 83 & 89 & 72 & 83 & 87 \\
& Pct & 14 & 100 & 100 & 43 & 86 & 100 \\
\addlinespace[0.3em]
Human & Avg\% & 75 & 69 & 76 & 75 & 71 & 70 \\
\bottomrule
\end{tabular}
}
\caption{Model performance on moral foundation classification across datasets. Acc\% shows balanced accuracy and Pct shows the corresponding percentile.}
\label{tab:accuracy}
\end{table}

The key insights are:

\textbf{AI Superiority over Humans.} AI models consistently outperformed humans across datasets, typically ranking in the top 25\% of annotators. Human performance averaged 67-76\%, while AI achieved 62-95\% depending on dataset.

\textbf{Similar Dataset Difficulty.} Average AI performance was similar across datasets: MFRC (83.7\%), eMFD (81.9\%), and MFTC (81.5\%), with negligible difference between the highest and lowest performing datasets.

\textbf{Model Strengths.} While all models outperformed human annotators, there is no strong winner among them. Claude appears to be a slight overall leader, scoring high consistently and achieving first or second-best results most of the time, with particular excellence in nuanced Care and Sanctity foundations. DeepSeek and Llama 4 perform similarly most of the time, but show a visible gap of 8-9\% behind Claude on MFRC in Sanctity and Fairness.

\textbf{Consistent Accuracy across Foundations.} All moral foundations achieved strong performance across datasets, with overall average accuracy exceeding 80\%: Care (85.1\%), Sanctity (84.3\%), Loyalty (81.6\%), Authority (80.9\%), and Fairness (80.0\%). Particularly high results for "Sanctity" are notable given its known difficulty to classify due to cultural contexts.

\subsection{False Positive vs. False Negative Analysis}

\begin{figure}[h!]
  \centering
  \includegraphics[width=0.99\linewidth]{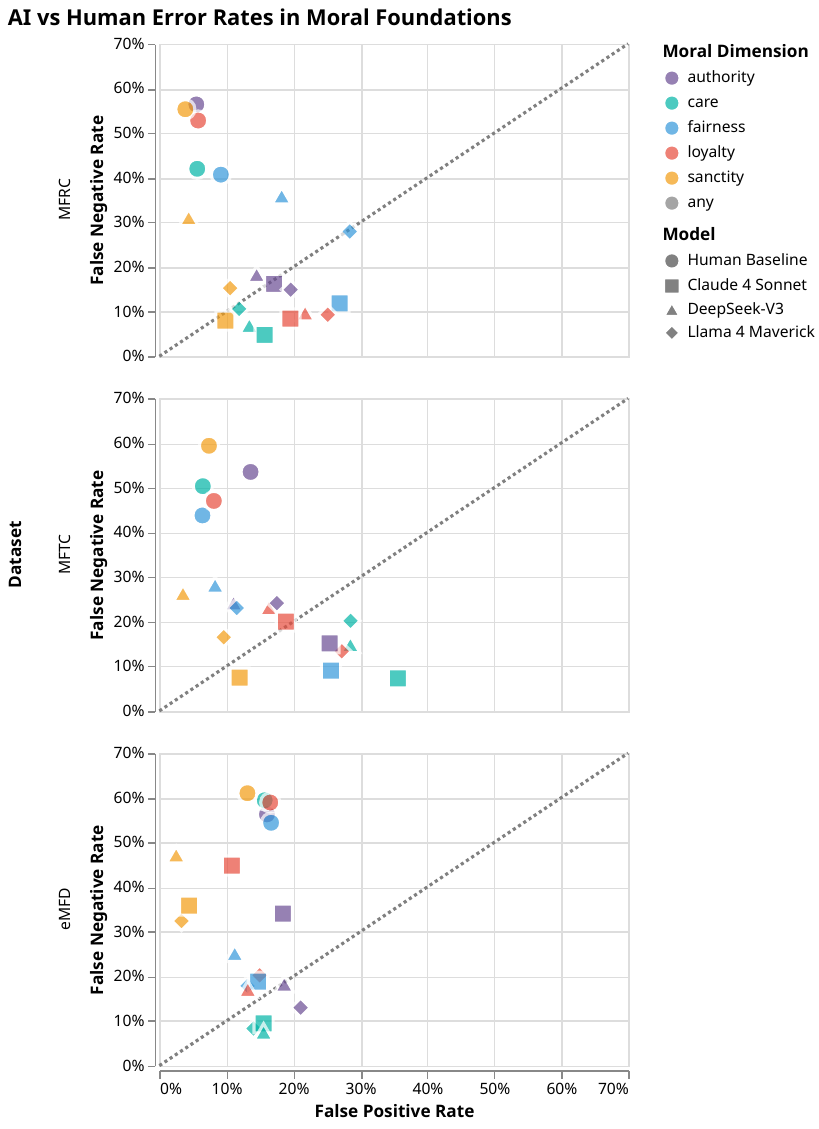}
    \caption{Error trade-offs in moral foundation detection. AI models (shapes) vs human baselines (circles) across datasets with colors denoting moral foundations. Diagonal lines indicate error balance (FPR = FNR).}
  \label{fig:fpr-fnr}
\end{figure}

\Cref{fig:fpr-fnr} and~\Cref{tab:fpr-fnr} compare false positive and false negative rates across models and foundations. The key findings are:

\textbf{AI-Human Error Trade-offs.} Across all foundations and datasets, AI models achieve $2−4\times$ lower false negative rates compared to humans (19.4\% vs 52.7\% on average), at the expense of slightly higher average false positive rates (16.1\% vs 10.1\%), overall remaining more balanced (as visually revealed in \Cref{fig:fpr-fnr}). This reveals opposing strategies: AI provides balanced detection, while humans classify more conservatively, systematically underdetect genuine moral signals.

\textbf{Foundation-Specific Patterns.} On average, the largest improvement in FNR appears in Care (40.8 point FNR versus humans), followed by Authority (35.6 points), Loyalty (34.5 points) and Sanctity (34.0 points). Fairness remains more challenging, with an average improvement of 24.2 points. 

\textbf{Human Conservatism Impact.} On MFRC and MFTC, annotated by a pool of experts, AI appears more balanced (average FNR-FPR AI gap of 1.4pp and 1.6pp, respectively) than in eMFD (gap of 8.7pp) crowd-annotated. This can be explained by the Bayesian method that estimates AI performance by seeking a consensus with a much higher number of conservative / biased annotators.

\begin{table}[h!]
  \centering
\resizebox{0.98\linewidth}{!}{
\begin{tabular}{@{}lrrrrrrrrrr@{}}
\toprule
\multicolumn{1}{c}{} & \multicolumn{10}{c}{Moral Dimensions} \\
\cmidrule(lr){2-11}
& \multicolumn{2}{c}{Authority} & \multicolumn{2}{c}{Care} & \multicolumn{2}{c}{Fairness} & \multicolumn{2}{c}{Loyalty} & \multicolumn{2}{c}{Sanctity} \\
\cmidrule(lr){2-3} \cmidrule(lr){4-5} \cmidrule(lr){6-7} \cmidrule(lr){8-9} \cmidrule(lr){10-11}
Model & FNR & FPR & FNR & FPR & FNR & FPR & FNR & FPR & FNR & FPR \\
\midrule
\multicolumn{11}{c}{\textbf{MFRC Dataset}} \\
\midrule
Claude Sonnet 4 & 16.5 & 17.2 & 5.3 & 15.2 & 12.3 & 27.2 & 8.7 & 19.5 & 7.9 & 9.5 \\
DeepSeek-V3 & 18.7 & 14.4 & 7.3 & 13.7 & 36.4 & 18.0 & 9.6 & 21.4 & 31.3 & 4.5 \\
Llama 4 Maverick & 14.6 & 20.0 & 10.8 & 11.4 & 28.1 & 28.8 & 9.7 & 24.8 & 15.2 & 10.5 \\
Human Baseline & 56.5 & 5.3 & 42.4 & 5.3 & 40.9 & 9.4 & 52.4 & 5.4 & 55.4 & 3.7 \\
\midrule
\multicolumn{11}{c}{\textbf{MFTC Dataset}} \\
\midrule
Claude Sonnet 4 & 14.7 & 25.3 & 7.2 & 35.8 & 9.3 & 25.8 & 20.1 & 18.7 & 7.8 & 12.2 \\
DeepSeek-V3 & 24.6 & 11.1 & 15.2 & 28.5 & 28.6 & 8.7 & 22.9 & 16.7 & 26.7 & 3.8 \\
Llama 4 Maverick & 23.9 & 18.0 & 19.9 & 28.2 & 22.8 & 11.5 & 13.3 & 27.7 & 16.1 & 9.2 \\
Human Baseline & 53.1 & 13.4 & 50.3 & 6.9 & 43.5 & 6.2 & 46.9 & 8.6 & 59.1 & 7.3 \\
\midrule
\multicolumn{11}{c}{\textbf{eMFD Dataset}} \\
\midrule
Claude Sonnet 4 & 33.6 & 18.4 & 9.1 & 15.2 & 19.3 & 15.0 & 44.9 & 10.6 & 36.1 & 4.1 \\
DeepSeek-V3 & 18.2 & 19.0 & 7.5 & 16.0 & 25.0 & 11.5 & 16.8 & 13.0 & 47.8 & 2.4 \\
Llama 4 Maverick & 12.6 & 20.7 & 7.9 & 14.1 & 18.2 & 13.4 & 19.9 & 14.8 & 32.2 & 3.8 \\
Human Baseline & 56.5 & 16.0 & 59.8 & 15.2 & 54.9 & 16.2 & 59.4 & 16.4 & 61.2 & 13.5 \\
\bottomrule
\end{tabular}
}
\caption{False positive and negative rates (\%) by model across moral foundations and datasets. Human baseline represents average performance across all annotators.}
\label{tab:fpr-fnr} 
\end{table}

\section{Case Study}

Human annotation, while considered the gold standard for supervised learning, may systematically underestimate moral content prevalence in text. As noted in annotator training guidelines, "Ambiguities present considerable challenges for human annotators who must strike an acceptable balance between exploiting often weak signals of moral sentiment while also avoiding unfounded speculation about authorial intent." Our results suggest AI models navigate this balance more effectively, achieving both higher recall rates and superior balanced accuracy compared to human annotators.

To verify this finding, we evaluated a selection of posts where AI predicted substantially more moral content than human annotators. An independent domain expert confirmed that AI was correct in the majority of cases, suggesting that human annotators systematically miss moral content that AI systems correctly identify. Full details of this validation study are provided in the supplementary material and code repository. The examples below represent cases where no human annotator detected moral language, false negatives in human annotation that AI correctly identified, demonstrating AI's superior sensitivity to subtle moral cues in text.

\begin{tcolorbox}[
colback=orange!5,
colframe=orange!50!black,
boxrule=1pt,
arc=4pt,
left=4pt,
right=4pt,
top=4pt,
bottom=4pt,
valign=top,
title={\textbf{Example: Claude Sonnet 4 Classification}},
fonttitle=\bfseries,
coltitle=white,
colbacktitle=orange!70!black
]
{\small
\textbf{User:} You are an expert in moral psychology, classifying text according to Haidt's theory (...)

Text: ``Anyone think Macron should dispose of the Algerian dictator? France did little to support the uprisings of 2010-2012.''
\\

\textbf{Assistant:}

\begin{verbatim}
{
"care/harm": true,
"fairness/cheating": true,
"loyalty/betrayal": false,
"authority/subversion": true,
"sanctity/degradation": false
}
\end{verbatim}
{Reasoning:}
\begin{itemize}
\item {Care/harm:} Concern for harm under ``dictator,'' suggests intervention to protect people
\item {Fairness:} Implies Algerians deserve better; criticizes France's abandonment
\item {Authority:} Direct challenge to political authority through regime change
\end{itemize}
}
\end{tcolorbox}

\begin{tcolorbox}[
colback=orange!5,
colframe=orange!50!black,
boxrule=1pt,
arc=4pt,
left=4pt,
right=4pt,
top=4pt,
bottom=4pt,
valign=top,
title={\textbf{Example: Claude Sonnet 4 Classification}},
fonttitle=\bfseries,
coltitle=white,
colbacktitle=orange!70!black
]
{\small
\textbf{User:} You are an expert in moral psychology, classifying text according to Haidt's theory (...)

Text: ``Victim blaming are we? Even if the therapist is her friend, she should've known not to tell her anything. It's immoral and extremely disrespectful, and being friends with his wife is no excuse.''
\\

\textbf{Assistant:}
\begin{verbatim}
{
"care/harm": true,
"fairness/cheating": true,
"loyalty/betrayal": true,
"authority/subversion": true,
"sanctity/degradation": true
}
\end{verbatim}
Reasoning:
\begin{itemize}
\item Care/harm: Concern for victim's wellbeing and harm from breach of confidentiality
\item Fairness: Views therapist's actions as unfair betrayal of professional ethics
\item Loyalty: Frames situation as betrayal of professional duties and patient trust
\item Authority: Challenges therapist's abuse of professional authority
\item Sanctity: Treats therapeutic confidentiality as sacred/inviolable
\end{itemize}
}
\end{tcolorbox}

\begin{tcolorbox}[
colback=orange!5,
colframe=orange!50!black,
boxrule=1pt,
arc=4pt,
left=4pt,
right=4pt,
top=4pt,
bottom=4pt,
valign=top,
title={\textbf{Example: Claude Sonnet 4 Classification}},
fonttitle=\bfseries,
coltitle=white,
colbacktitle=orange!70!black
]
{\small
\textbf{User:} You are an expert in moral psychology, classifying text according to Haidt's theory (...)

Text: ``If you care about this cat so much and it has been a WEEK since you have seen her, WHY HAVE YOU NOT CALLED THE POLICE AND REPORTED THE THEFT??? Once you do all that, you need to cut contact with your "boyfriend". I would've had my cat back day of or would've filed the theft immediately. This is pushing ESH because of how long you've waited for YOUR CAT.''
\\

\textbf{Assistant:}
\begin{verbatim}
{
"care/harm": true,
"fairness/cheating": true,
"loyalty/betrayal": false,
"authority/subversion": true,
"sanctity/degradation": false
}
\end{verbatim}
Reasoning:
\begin{itemize}
\item Care/harm: Strong concern for cat's welfare and emotional distress from separation
\item Fairness: Views theft as clear injustice requiring legal consequences
\item Authority: Appeals to police authority to enforce justice and restore order
\end{itemize}
}
\end{tcolorbox}

\section*{Conclusion}

This work presents the most comprehensive evaluation to date of moral foundation detection capabilities in state-of-the-art large language models, analyzing 1M+ queries across 250K+ annotations from three established corpora. By employing Bayesian annotation modeling, we moved beyond deterministic ground-truth assumptions to explicitly account for annotator disagreement in moral foundation classification.

Our key findings demonstrate that AI models consistently outperform human annotators in moral foundation detection, typically ranking in the top 25\% across all datasets. Critically, AI models achieve $2-4\times$  lower false negative rates, while accepting slightly higher false positive rates, revealing fundamentally different error strategies. Although humans often miss genuine moral foundations due to biases, AI models provide more balanced detection of moral content.


These findings have important implications for deploying LLMs in moral foundation analysis. The superior recall capabilities make AI models valuable for detecting moral foundations that humans might overlook, though slightly higher false positive rates require careful calibration for specific applications. Our uncertainty-aware evaluation framework provides a foundation for more nuanced assessment of AI moral foundation detection that accounts for inherent subjectivity rather than treating classification as deterministic.

Future work should focus on improving model calibration for moral foundation detection and extending evaluation to other moral frameworks, as well as exploring connections to other datasets such as anticipatory discourses~\cite{landowskaWhatTwitterData2023b,skorskiMappingTechnologicalFutures2025}.

\section*{Acknowledgements}

The authors thank the people and institutions who supported this research: Murmuras GmbH\footnote{\url{https://murmuras.com/}} for providing access to state-of-the-art large language models' APIs, University of Luxembourg for financial support and computing infrastructure, Kozminski University and Polish Ministry of Science for supporting participation in the "Top 1000 Innovators of Poland in Silicon Valley" program, Pierre Pailler from Virgule for the opportunity to present this research to the general public~\cite{paillerpierreLIAMeilleureQue2025}, and the anonymous reviewers of the UncertainNLP workshop for their valuable feedback.

\section*{Limitations}

\paragraph{Fixed Effects Modeling Limitations.} While our Bayesian framework supports demographic fixed effects to account for systematic annotator differences, we did not employ these due to limited demographic metadata (only the crowd-sourced eMFD corpus contains sufficient demographic data). This somewhat constraints our ability to model known sources of variation in moral foundation detection across annotator groups.

\paragraph{Content Moderation Limitations.} Content moderation systems may introduce systematic bias by refusing to process morally relevant content. Azure OpenAI rejected ~10\% of Twitter data citing improper content, while Anthropic's model questioned the factual accuracy of referring to an "Algerian dictator" in our Macron example. After excluding OpenAI, content moderation affected less than 0.5\% of our final evaluation dataset.

\paragraph{Data Availability Constraints.} The three corpora span different periods (2016-2022) and textual domains, reflecting constraints of available annotated datasets rather than ideal experimental design. These temporal and domain variations may somewhat limit the generalizability of the findings.

\paragraph{Annotation Protocol Heterogeneity.} The three corpora employed different annotation protocols—varying in annotator training, guidelines, and expertise levels—which may introduce inconsistencies in the ground truth labels. The MFTC used trained experts while eMFD relied on crowd workers, potentially creating systematic differences in annotation quality and interpretation that our Bayesian model may not fully reconcile.

\paragraph{Prompt Sensitivity Limitations.} While we deliberately avoided extensive prompt engineering to reflect typical usage conditions, LLM performance on moral classification tasks is known to be sensitive to prompt phrasing and structure (see~\cite{skorskiMoralGapLarge2025}). Our single-prompt approach yields balanced results suitable for general-purpose evaluation, though alternative prompt strategies (e.g., chain-of-thought, few-shot examples, explicit instructions emphasizing sensitivity or specificity) could shift the precision-recall tradeoff. Such variations might improve performance on specific moral concepts or optimize for particular error patterns, but would deviate from typical deployment conditions we aimed to assess.

\paragraph{Low-Resource Language and Cultural Bias Limitations.} Our evaluation is limited to English-language datasets, as annotated moral reasoning datasets in low-resource languages are not yet available. This gap somewhat limits cross-cultural validation of our findings, though annotator diversity—from trained experts to crowdsourced contributors—helps mitigate individual cultural biases within English contexts. As the field develops multilingual moral foundation datasets, our framework can readily extend to cross-linguistic evaluation.

\paragraph{Theoretical Scope Limitations.} Our exclusive focus on Moral Foundation Theory, while practical and widely-used for large-scale analysis, represents a significant theoretical limitation. Alternative moral and ethical theories (e.g., virtue ethics~\cite{culhamVirtueEthicsMoral2024a}, deontological ethics~\cite{gawronskiWhatMakesMoral2016} or care ethics~\cite{deakValuingCareProtects2016}) offer different perspectives on morals that may capture dimensions not encompassed by MFT's five foundations. Furthermore, the binary classification framework fails to capture the intensity, valence, or contextual nuance of moral expressions. However, MFT remains best suited for large-scale computational analysis given its established usage, data availability, and continued popularity in computational linguistics research. Future work should explore hybrid approaches that combine MFT with complementary frameworks like virtue ethics or care ethics to provide more comprehensive moral analysis.




\bibliography{citations}

@article{graham2009liberals,
  title={Liberals and conservatives rely on different sets of moral foundations},
  author={Graham, Jesse and Haidt, Jonathan and Nosek, Brian A},
  journal={Journal of personality and social psychology},
  volume={96},
  number={5},
  pages={1029--1046},
  year={2009},
  publisher={American Psychological Association},
    doi = {10.1037/a0015141},
}

@article{feinberg2013moral,
  title={The Moral Roots of Environmental Attitudes},
  author={Feinberg, Matthew and Willer, Robb},
  journal={Psychological Science},
  volume={24},
  number={1},
  pages={56--62},
  year={2013}
}

@inproceedings{nguyen2022mapping,
  title={Mapping Topics in 100,000 Real-Life Moral Dilemmas},
  author={Nguyen, Tuan Dung and Lyall, Georgina and Tran, Alasdair and Shin, Minkyoung and Carroll, Nicholas G and Klein, Colin and Xie, Lexing},
  booktitle={Proceedings of the International AAAI Conference on Web and Social Media},
  volume={16},
  pages={699--710},
  year={2022}
}

@book{haidt2012righteous,
  title = {The {{Righteous Mind}}: {{Why Good People Are Divided}} by {{Politics}} and {{Religion}}},
  shorttitle = {The {{Righteous Mind}}},
  author = {Haidt, Jonathan},
  year = {2012},
  month = mar,
  publisher = {Knopf Doubleday Publishing Group},
  googlebooks = {ItuzJhbcpMIC},
  isbn = {978-0-307-90703-5},
  langid = {english}
}

@article{dawidMaximumLikelihoodEstimation1979,
  title = {Maximum {{Likelihood Estimation}} of {{Observer Error-Rates Using}} the {{EM Algorithm}}},
  author = {Dawid, A. P. and Skene, A. M.},
  year = {1979},
  journal = {Applied Statistics},
  volume = {28},
  number = {1},
  eprint = {10.2307/2346806},
  eprinttype = {jstor},
  pages = {20},
  issn = {00359254},
  doi = {10.2307/2346806},
  url = {https://www.jstor.org/stable/10.2307/2346806?origin=crossref},
  urldate = {2025-09-30}
}

@article{graham2013moral,
  title={Moral Foundations Theory: The Pragmatic Validity of Moral Pluralism},
  author={Graham, Jesse and Haidt, Jonathan and Koleva, Sena and Motyl, Matt and Iyer, Ravi and Wojcik, Sean P and Ditto, Peter H},
  journal={Advances in Experimental Social Psychology},
  volume={47},
  pages={55--130},
  year={2013},
  publisher={Elsevier}
}

@inproceedings{paunAggregatingLearningMultiple2021,
  title = {Aggregating and {{Learning}} from {{Multiple Annotators}}},
  booktitle = {Proceedings of the 16th {{Conference}} of the {{European Chapter}} of the {{Association}} for {{Computational Linguistics}}: {{Tutorial Abstracts}}},
  author = {Paun, Silviu and Simpson, Edwin},
  year = {2021},
  pages = {6--9},
  publisher = {Association for Computational Linguistics},
  address = {online},
  doi = {10.18653/v1/2021.eacl-tutorials.2},
  url = {https://aclanthology.org/2021.eacl-tutorials.2},
  urldate = {2025-08-03},
  langid = {english}
}

@article{nguyenMeasuringMoralDimensions2024a,
  title = {Measuring {{Moral Dimensions}} in {{Social Media}} with {{Mformer}}},
  author = {Nguyen, Tuan Dung and Chen, Ziyu and Carroll, Nicholas George and Tran, Alasdair and Klein, Colin and Xie, Lexing},
  year = {2024},
  month = may,
  journal = {Proceedings of the International AAAI Conference on Web and Social Media},
  volume = {18},
  pages = {1134--1147},
  issn = {2334-0770, 2162-3449},
  doi = {10.1609/icwsm.v18i1.31378},
  url = {https://ojs.aaai.org/index.php/ICWSM/article/view/31378},
  urldate = {2025-05-29}
}

@article{zangariSurveyMoralFoundation2025,
  title = {A Survey on Moral Foundation Theory and Pre-Trained Language Models: Current Advances and Challenges},
  shorttitle = {A Survey on Moral Foundation Theory and Pre-Trained Language Models},
  author = {Zangari, Lorenzo and Greco, Candida Maria and Picca, Davide and Tagarelli, Andrea},
  year = {2025},
  month = mar,
  journal = {AI \& SOCIETY},
  issn = {0951-5666, 1435-5655},
  doi = {10.1007/s00146-025-02225-w},
  url = {https://link.springer.com/10.1007/s00146-025-02225-w},
  urldate = {2025-05-29},
  langid = {english}
}

@article{bullaLargeLanguageModels2025,
  title = {Large {{Language Models}} Meet Moral Values: {{A}} Comprehensive Assessment of Moral Abilities},
  shorttitle = {Large {{Language Models}} Meet Moral Values},
  author = {Bulla, Luana and De Giorgis, Stefano and Mongiov{\`i}, Misael and Gangemi, Aldo},
  year = {2025},
  month = mar,
  journal = {Computers in Human Behavior Reports},
  volume = {17},
  pages = {100609},
  issn = {24519588},
  doi = {10.1016/j.chbr.2025.100609},
  url = {https://linkinghub.elsevier.com/retrieve/pii/S2451958825000247},
  urldate = {2025-05-29},
  langid = {english}
}

@inproceedings{preniqiMoralBERTFineTunedLanguage2024,
  title = {{{MoralBERT}}: {{A Fine-Tuned Language Model}} for {{Capturing Moral Values}} in {{Social Discussions}}},
  shorttitle = {{{MoralBERT}}},
  booktitle = {Proceedings of the 2024 {{International Conference}} on {{Information Technology}} for {{Social Good}}},
  author = {Preniqi, Vjosa and Ghinassi, Iacopo and Ive, Julia and Saitis, Charalampos and Kalimeri, Kyriaki},
  year = {2024},
  month = sep,
  pages = {433--442},
  publisher = {ACM},
  address = {Bremen Germany},
  doi = {10.1145/3677525.3678694},
  url = {https://dl.acm.org/doi/10.1145/3677525.3678694},
  urldate = {2025-05-29},
  isbn = {9798400710940},
  langid = {english}
}

@inproceedings{zangari-etal-2025-me2,
  title = {{{ME2-BERT}}: {{Are}} Events and Emotions What You Need for Moral Foundation Prediction?},
  booktitle = {Proceedings of the 31st International Conference on Computational Linguistics},
  author = {Zangari, Lorenzo and Greco, Candida M. and Picca, Davide and Tagarelli, Andrea},
  editor = {Rambow, Owen and Wanner, Leo and Apidianaki, Marianna and {Al-Khalifa}, Hend and Eugenio, Barbara Di and Schockaert, Steven},
  year = {2025},
  month = jan,
  pages = {9516--9532},
  publisher = {Association for Computational Linguistics},
  address = {Abu Dhabi, UAE},
  url = {https://aclanthology.org/2025.coling-main.638/}
}

@article{hooverMoralFoundationsTwitter2020,
  title = {Moral {{Foundations Twitter Corpus}}: {{A Collection}} of 35k {{Tweets Annotated}} for {{Moral Sentiment}}},
  shorttitle = {Moral {{Foundations Twitter Corpus}}},
  author = {Hoover, Joe and {Portillo-Wightman}, Gwenyth and Yeh, Leigh and Havaldar, Shreya and Davani, Aida Mostafazadeh and Lin, Ying and Kennedy, Brendan and Atari, Mohammad and Kamel, Zahra and Mendlen, Madelyn and Moreno, Gabriela and Park, Christina and Chang, Tingyee E. and Chin, Jenna and Leong, Christian and Leung, Jun Yen and Mirinjian, Arineh and Dehghani, Morteza},
  year = {2020},
  month = nov,
  journal = {Social Psychological and Personality Science},
  volume = {11},
  number = {8},
  pages = {1057--1071},
  issn = {1948-5506, 1948-5514},
  doi = {10.1177/1948550619876629},
  url = {https://journals.sagepub.com/doi/10.1177/1948550619876629},
  urldate = {2025-08-11},
  langid = {english}
}

@article{skorskiMappingTechnologicalFutures2025,
  title = {Mapping Technological Futures: Anticipatory Discourse through Text Mining},
  shorttitle = {Mapping Technological Futures},
  author = {Sk{\'o}rski, Maciej and Landowska, Alina and Rajda, Krzysztof},
  year = {2025},
  month = jun,
  journal = {Humanities and Social Sciences Communications},
  volume = {12},
  number = {1},
  pages = {899},
  issn = {2662-9992},
  doi = {10.1057/s41599-025-05083-5},
  url = {https://www.nature.com/articles/s41599-025-05083-5},
  urldate = {2025-09-22},
  langid = {english}
}

@misc{landowskaWhatTwitterData2023b,
  title = {What {{Twitter Data Tell Us}} about the {{Future}}?},
  author = {Landowska, Alina and Robak, Marek and Skorski, Maciej},
  year = {2023},
  month = jul,
  number = {arXiv:2308.02035},
  eprint = {2308.02035},
  primaryclass = {cs},
  publisher = {arXiv},
  doi = {10.48550/arXiv.2308.02035},
  url = {http://arxiv.org/abs/2308.02035},
  urldate = {2025-09-22},
  archiveprefix = {arXiv}
}

@article{skorskiBayesianEvaluationLLMs2025,
  title = {Bayesian {{Evaluation}} of {{LLMs}}' {{Moral Values Understanding}}},
  author = {Sk{\'o}rski, Maciej},
  year = {2025},
  publisher = {OSF},
  doi = {10.17605/OSF.IO/TPZAU},
  url = {https://osf.io/tpzau/},
  urldate = {2025-09-22},
  collaborator = {{Center for Open Science}},
note = {Supplementary materials},
  copyright = {Creative Commons Attribution 4.0 International},

}

@article{paillerpierreLIAMeilleureQue2025,
  title = {{L'IA meilleure que l'humain sur les questions morales, selon une {\'e}tude de l'Uni?}},
  author = {Pailler, Pierre},
  year = {2025},
  month = sep,
  journal = {Virgule},
  url = {https://www.virgule.lu/luxembourg/l-ia-meilleure-que-l-humain-sur-les-questions-morales-selon-une-etude-de-l-uni/90599367.html},
  langid = {french}
}

@article{deakValuingCareProtects2016,
  title = {Valuing {{Care Protects Religiosity}} from the {{Antisocial Consequences}} of {{Impersonal Deontology}}},
  author = {Deak, Csilla and Saroglou, Vassilis},
  year = {2016},
  month = dec,
  journal = {Journal of Empirical Theology},
  volume = {29},
  number = {2},
  pages = {171--189},
  issn = {0922-2936, 1570-9256},
  doi = {10.1163/15709256-12341339},
  url = {https://brill.com/view/journals/jet/29/2/article-p171_4.xml},
  urldate = {2025-09-22}
}

@article{gawronskiWhatMakesMoral2016,
  title = {What Makes Moral Dilemma Judgments ``Utilitarian'' or ``Deontological''?},
  author = {Gawronski, Bertram and Beer, Jennifer S.},
  year = {2016},
  month = oct,
  journal = {Social Neuroscience},
  pages = {1--7},
  issn = {1747-0919, 1747-0927},
  doi = {10.1080/17470919.2016.1248787},
  url = {https://www.tandfonline.com/doi/full/10.1080/17470919.2016.1248787},
  urldate = {2025-10-07},
  langid = {english}
}

@misc{tensorflow2015-whitepaper,
title={ {TensorFlow}: Large-Scale Machine Learning on Heterogeneous Systems},
url={https://www.tensorflow.org/},
note={Software available from tensorflow.org},
author={
    Mart\'{i}n~Abadi and
    Ashish~Agarwal and
    Paul~Barham and
    Eugene~Brevdo and
    Zhifeng~Chen and
    Craig~Citro and
    Greg~S.~Corrado and
    Andy~Davis and
    Jeffrey~Dean and
    Matthieu~Devin and
    Sanjay~Ghemawat and
    Ian~Goodfellow and
    Andrew~Harp and
    Geoffrey~Irving and
    Michael~Isard and
    Yangqing Jia and
    Rafal~Jozefowicz and
    Lukasz~Kaiser and
    Manjunath~Kudlur and
    Josh~Levenberg and
    Dandelion~Man\'{e} and
    Rajat~Monga and
    Sherry~Moore and
    Derek~Murray and
    Chris~Olah and
    Mike~Schuster and
    Jonathon~Shlens and
    Benoit~Steiner and
    Ilya~Sutskever and
    Kunal~Talwar and
    Paul~Tucker and
    Vincent~Vanhoucke and
    Vijay~Vasudevan and
    Fernanda~Vi\'{e}gas and
    Oriol~Vinyals and
    Pete~Warden and
    Martin~Wattenberg and
    Martin~Wicke and
    Yuan~Yu and
    Xiaoqiang~Zheng},
  year={2015},
}

@article{culhamVirtueEthicsMoral2024a,
  title = {Virtue Ethics and Moral Foundation Theory Applied to Business Ethics Education},
  author = {Culham, Tom E. and Major, Richard J. and Shivhare, Neha},
  year = {2024},
  month = apr,
  journal = {International Journal of Ethics Education},
  volume = {9},
  number = {1},
  pages = {139--176},
  issn = {2363-9997, 2364-0006},
  doi = {10.1007/s40889-023-00181-x},
  url = {https://link.springer.com/10.1007/s40889-023-00181-x},
  urldate = {2025-09-22},
  langid = {english}
}

@article{skorskiMoralGapLarge2025,
  title = {The {{Moral Gap}} of {{Large Language Models}}},
  author = {Skorski, Maciej and Landowska, Alina},
  year = {2025},
  eprint = {2507.18523},
  primaryclass = {cs},
  doi = {10.13140/RG.2.2.26221.70880},
  url = {http://arxiv.org/abs/2507.18523},
  urldate = {2025-08-19},
  archiveprefix = {arXiv}
}

@inproceedings{royAnalysisNuancedStances2021,
  title = {Analysis of {{Nuanced Stances}} and {{Sentiment Towards Entities}} of {{US Politicians}} through the {{Lens}} of {{Moral Foundation Theory}}},
  booktitle = {Proceedings of the {{Ninth International Workshop}} on {{Natural Language Processing}} for {{Social Media}}},
  author = {Roy, Shamik and Goldwasser, Dan},
  year = {2021},
  pages = {1--13},
  publisher = {Association for Computational Linguistics},
  address = {Online},
  doi = {10.18653/v1/2021.socialnlp-1.1},
  url = {https://www.aclweb.org/anthology/2021.socialnlp-1.1},
  urldate = {2025-08-16},
  langid = {english}
}

@article{hoppExtendedMoralFoundations2021,
  title = {The Extended {{Moral Foundations Dictionary}} ({{eMFD}}): {{Development}} and Applications of a Crowd-Sourced Approach to Extracting Moral Intuitions from Text},
  shorttitle = {The Extended {{Moral Foundations Dictionary}} ({{eMFD}})},
  author = {Hopp, Frederic R. and Fisher, Jacob T. and Cornell, Devin and Huskey, Richard and Weber, Ren{\'e}},
  year = {2021},
  month = feb,
  journal = {Behavior Research Methods},
  volume = {53},
  number = {1},
  pages = {232--246},
  issn = {1554-3528},
  doi = {10.3758/s13428-020-01433-0},
  url = {https://link.springer.com/10.3758/s13428-020-01433-0},
  urldate = {2025-05-29},
  langid = {english}
}

@misc{tragerMoralFoundationsReddit2022,
  title = {The {{Moral Foundations Reddit Corpus}}},
  author = {Trager, Jackson and Ziabari, Alireza S. and Davani, Aida Mostafazadeh and Golazizian, Preni and {Karimi-Malekabadi}, Farzan and Omrani, Ali and Li, Zhihe and Kennedy, Brendan and Reimer, Nils Karl and Reyes, Melissa and Cheng, Kelsey and Wei, Mellow and Merrifield, Christina and Khosravi, Arta and Alvarez, Evans and Dehghani, Morteza},
  year = {2022},
  publisher = {arXiv},
  doi = {10.48550/ARXIV.2208.05545},
  url = {https://arxiv.org/abs/2208.05545},
  urldate = {2025-05-29},
  copyright = {Creative Commons Attribution 4.0 International}
}

\appendix

\section{AI Prompt}\label{sec:prompt}

For estimating moral competence under typical usage conditions, we deliberately avoided prompt engineering techniques that might bias models toward higher recall or precision (the work of \cite{skorskiMoralGapLarge2025} demonstrated that LLMs are somewhat sensitive in that regard). We tested prompt variants that explicitly request step-by-step reasoning, without noticing significant differences.

\begin{tcolorbox}[
    colback=blue!5,
    colframe=blue!50!black,
    boxrule=1.5pt,
    arc=4pt,
    left=3pt,
    right=3pt,
    top=3pt,
    bottom=3pt,
    title={\textbf{Moral Foundations Theory Classification}},
    fonttitle=\bfseries,
    coltitle=white,
    colbacktitle=blue!75!black
]

{\small You are an expert in moral psychology, classifying text according to Haidt's theory.

For each moral foundation, mark \texttt{true} if moral values from that foundation are expressed in the text, \texttt{false} if not expressed.

\vspace{4pt}

\textbf{Answer only with a valid JSON in this format:}

\begin{verbatim}
{
    "care/harm": [true / false],
    "fairness/cheating": [true / false], 
    "loyalty/betrayal": [true / false],
    "authority/subversion": [true / false],
    "sanctity/degradation": [true / false],
    "reasoning": [summary of reasoning],
}
\end{verbatim}

Provide step-by-step reasoning.
}
\end{tcolorbox}

\section{TensorFlow Model Implementation}
\label{sec:implementation}

.
Our Bayesian aggregation uses sparse tensor operations for efficient likelihood computation across annotator-item pairs. The implementation leverages TensorFlow's sparse lookup operations to handle high-dimensional confusion matrices, enabling scalable inference on GPU hardware. The core algorithm computes marginal likelihoods over latent true labels by efficiently aggregating log-probabilities from annotator-specific confusion matrices, exploiting sparsity in the annotation pattern (most annotators do not label all items). The optimization uses gradient ascent on the log-posterior, jointly estimating class prevalences and per-annotator competence parameters. Details are available in our code repository~\cite{skorskiBayesianEvaluationLLMs2025}.

\begin{lstlisting}[language=Python, caption=TensorFlow implementation, basicstyle=\footnotesize, breaklines=true]

def log_p(pi_logits, theta_logits, annot_ids):
   """
   Implements the log-likelihood computation for a Dawid-Skene competence model, 
   estimating both class prevalences and annotator confusion matrices.
   
   Parameters:
   -----------
   pi_logits : tf.Variable, shape [K]
       Logits for class prevalence distribution pi (before softmax normalization)
   theta_logits : tf.Variable, shape [J, K, K] 
       Logits for annotator confusion matrices theta (before softmax normalization)
       theta[j,i,k] = P(annotator j labels class k | true class i)
   annot_ids : tf.SparseTensor, shape [N, J*K]
       Sparse tensor encoding annotation observations where:
       - indices: (item, annotator) pairs
       - values: observed classes encoded for efficient embedding lookup
       
   Returns:
   --------
   tf.Tensor : scalar
       Log-likelihood = log P(annotations | pi, theta) + log P(pi) + log P(theta)
       Combines data likelihood with Dirichlet priors on pi and theta
       
   Mathematical formulation:
   ------------------------
   log P(annotations) = sum_i log sum_k pi_k prod_j theta_jk,y_ij + log P(pi) + sum_j log P(theta_j)
   where y_ij is the annotation by annotator j on item i
   """
    log_pi = tf.nn.log_softmax(pi_logits)
    log_theta = tf.nn.log_softmax(theta_logits, axis=-1) # [annotator x true class x pred class]
    pi = tf.math.exp(log_pi)  # [true class]
    theta = tf.math.exp(log_theta)
    log_theta = tf.transpose(log_theta,[0,2,1]) 
    log_theta = tf.reshape(log_theta, (J*K,K)) # [annotator * true class, x pred class]
    log_p = tf.nn.embedding_lookup_sparse(log_theta, annot_ids, sp_weights=None, combiner='sum')  # [items x true class]
    log_p += log_pi[None, :]
    log_p = tf.reduce_logsumexp(log_p, axis=1) # [items]
    log_p = tf.reduce_sum(log_p)
    
    log_p += class_prior.log_prob(pi)
    log_p += tf.reduce_sum(confusion_prior.log_prob(theta))
    return log_p

optimizer = tf.optimizers.Adam(1e-2, )
max_iter = 2000

@tf.function()
def train_step(pi_logits, theta_logits, annot_ids):
    with tf.GradientTape() as tape:
        loss = -log_p(pi_logits, theta_logits, annot_ids)
    gradients = tape.gradient(loss, [pi_logits, theta_logits])
    optimizer.apply_gradients(zip(gradients, [pi_logits, theta_logits]))
    return loss

@tf.function()
def train(pi_logits, theta_logits, annot_ids, max_iter=tf.constant(1)):
    print("tracing")
    for i in tf.range(max_iter):
        loss = train_step(pi_logits, theta_logits, annot_ids )   

# get competences from logits
theta = tf.nn.softmax(theta_logits, axis=-1)
competences = tf.reduce_sum(tf.ones(K)*1.0/K * tf.linalg.diag_part(theta), axis=1).numpy()

\end{lstlisting}

\section{Content Moderation Examples}

Social media is full of expressive posts that occasionally may not be evaluated by AI providers who implement strict internal safety mechanisms, limiting research capabilities on real-world content. For this reason, we did not include the results of OpenAI. One rejected example is shown below.

\begin{tcolorbox}[
colback=red!5,
colframe=red!50!black,
boxrule=1pt,
arc=4pt,
left=2pt,
right=2pt,
top=5pt,
bottom=5pt,
title={\textbf{Example: Azure OpenAI Content Filtering}},
fonttitle=\bfseries,
coltitle=white,
colbacktitle=red!70!black
]
{\small
\textbf{Input:} ``Stop racist black thugs \& their Marxist masters terror campaign. \#LockThemUp \#NoBail \#alllivesmatter \#PoliceLivesMatter''
\\
\textbf{Response:} Err. 400: Content management policy violation
\\
\textbf{Filter Results:}
\begin{itemize}[noitemsep,topsep=0pt]
\item \textbf{Hate:} Filtered (High severity)
\item \textbf{Violence:} Filtered (Medium severity)
\item \textbf{Sexual/Self-harm:} Safe
\end{itemize}
}
\end{tcolorbox}

\section{Moral Foundations Color Palette}
\label{app:color-palette}

We use the colorblind palette from~\cite{skorskiMoralGapLarge2025}, transitioning from individualistic (cooler) to collectivistic (warmer) foundations
\begin{figure}[h!]
\centering
\begin{tikzpicture}
\def\spacing{0.8}
\foreach \x/\colorname/\label in {
   0/care/Care,
   \spacing/fairness/Fairness,
   2*\spacing/loyalty/Loyalty,
   3*\spacing/authority/Authority,
   4*\spacing/sanctity/Sanctity
} {
  \fill[\colorname] (\x+0.6,0) -- (\x,0.15) -- (\x+0.1,0) -- (\x,-0.15) -- cycle;
  \node[above] at (\x+0.3,0.25) {\tiny\label};
}
\node[below] at (0,-0.4) {\small Individualism};
\node[below] at (4*\spacing+0.6,-0.4) {\small Collectivism};
\end{tikzpicture}
\caption{Moral colors: individualism to collectivism}
\label{fig:moral-palette}
\end{figure}

\end{document}